\documentclass[cameraready]{Interspeech}

\title{Vividh-ASR: A Complexity-Tiered Benchmark and Optimization Dynamics for Robust Indic Speech Recognition}

\author[orcid=0009-0000-8240-0099,equalcontribution]{Kush}{Juvekar}
\author[orcid=0000-0003-2402-5272,equalcontribution]{Kavya}{Manohar}
\author[orcid=0009-0000-9326-028X]{Aditya Srinivas} {Menon} 
\author[orcid=0009-0007-3192-4563]{Arghya}{Bhattacharya}
\author[orcid=0009-0003-0730-6897]{Kumarmanas}{Nethil}

\address{
    Adalat AI, India
}
\email{\{kush,kavya,dia,arghya,manas\}@adalat.ai}
\keywords{speech recognition, curriculum learning, Indic languages, fine-tuning}

\usepackage{comment}
\usepackage{tikz}
\usepackage{subfigure}
\usetikzlibrary{shapes.geometric, arrows, positioning, calc, shadows.blur}
\usepackage{pgfplots}
\pgfplotsset{compat=1.18}
\usepackage{tabularx}
\usepackage{booktabs}

\begin{document}
\maketitle

% ================================================================
% ABSTRACT
% ================================================================
\begin{abstract}

Fine-tuning multilingual ASR models like Whisper for low-resource languages often improves read speech but degrades spontaneous audio performance. To diagnose this mismatch, we introduce Vividh-ASR, a complexity-stratified benchmark for Hindi and Malayalam across four tiers: studio, broadcast, spontaneous, and synthetic noise. Through a controlled study of learning-rate timing and curriculum ordering, we find that early large parameter updates improve global WER by $\sim$12 absolute points, while a hard-to-easy curriculum adds gains for spontaneous speech. These findings motivate reverse multi-stage fine-tuning (R-MFT), a training recipe that enables a parameter-efficient 244M Whisper model to match or exceed conventionally fine-tuned 769M counterparts. Representational analysis via CKA and SVD reveals effective schedules concentrate adaptation in the decoder, preserving the pre-trained encoder’s acoustic geometry. We release the benchmark and models.
\end{abstract}

% ================================================================
% 1. INTRODUCTION
% ================================================================
\section{Introduction}
\label{sec:intro}

Large-scale weakly supervised pre-training has brought automatic speech recognition (ASR) close to human performance in high-resource languages~\cite{radford2023robust, conneau2021xlsr}. However, zero-shot word error rates (WER) for many Indic languages often exceed 100\%. Fine-tuned models such as IndicWhisper~\cite{bhogale23_interspeech} reduce this gap, but are trained primarily on studio-recorded read speech. Consequently, they perform well on clean audio yet degrade sharply on spontaneous conversational speech---a failure mode we term \textit{studio-bias}.

Two conventions dominate current fine-tuning practice for models like Whisper. First, conservative learning rates ($1\mathrm{e}^{-5}$) are commonly used to avoid catastrophic forgetting~\cite{lodagala_whisper_finetune_hyperparams,kirkpatrick2017ewc} of the multilingual prior. Second, when curricula are employed, training typically follows an easy-to-hard progression~\cite{bengio2009curriculum} that gradually introduces noisier and more spontaneous speech. These heuristics implicitly assume that the pre-trained encoder already contains the acoustic structure necessary for low-resource target languages. In practice, however, adapting to the phonotactics and articulation of spontaneous Dravidian and Indo-Aryan speech, which dominates usecases like real-world courtroom dictation \cite{nethil25_interspeech}, requires substantial early plasticity.

Rather than relying on data scaling, we argue that adaptation efficiency is governed by two interacting factors: the \textit{timing} of large parameter updates and the \textit{ordering} of acoustic complexity. We systematically decouple these dynamics through a controlled $2\times2$ factorial study, isolating the effects of learning-rate timing and curriculum direction while holding the training data, model architecture, and optimizer configuration constant. Our findings challenge the status quo: we demonstrate that reversing standard heuristics---applying high-magnitude updates initially on the hardest data---yields drastic WER improvements on identical data distributions, while a conservative initialization traps the model in a sub-optimal basin.

To better understand these dynamics, we analyze how different optimization schedules reshape internal model representations using centered kernel alignment (CKA)~\cite{kornblith2019similarity} and singular value decomposition (SVD). This suggests that effective schedules concentrate most adaptation within the decoder while largely preserving the encoder’s pre-trained acoustic geometry. We make two primary contributions:

\begin{enumerate}
    \item \textbf{Vividh-ASR Benchmark.} A diagnostic benchmark for Hindi and Malayalam that stratifies evaluation by acoustic complexity: studio (Tier~A), broadcast (Tier~B), spontaneous (Tier~C), and synthetic noise (Tier~D). Unlike domain-based benchmarks, this structure isolates precisely where models fail along the complexity axis.

    \item \textbf{Reverse multi-stage fine-tuning (R-MFT).} A training recipe that pairs spontaneous-first data ordering with high initial learning rates. We release a parameter-efficient 244M Whisper model trained with this recipe, demonstrating its efficacy on spontaneous Indic speech.
\end{enumerate}

% [REVISED: Moved figure to after contributions for better flow]
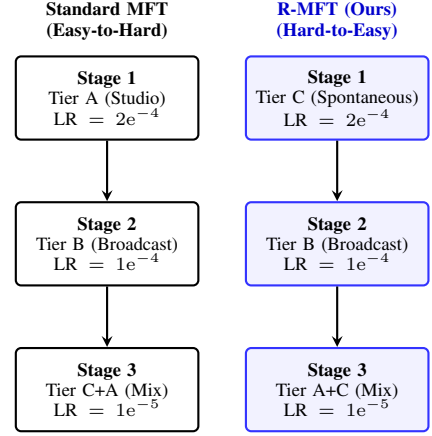
\begin{figure}[t]
\centering
\begin{tikzpicture}[
    node distance=0.8cm and 0.4cm,
    stage/.style={rectangle, rounded corners=2pt, draw=black, thick, fill=white, text width=2.2cm, align=center, minimum height=1.1cm, font=\scriptsize},
    revstage/.style={rectangle, rounded corners=2pt, draw=blue!80, thick, fill=blue!5, text width=2.2cm, align=center, minimum height=1.1cm, font=\scriptsize},
    arrow/.style={-stealth, thick},
    header/.style={font=\scriptsize\bfseries, align=center, text width=2.8cm}
]
% --- Standard MFT (Left Side) ---
\node (s1_std) [stage] {\textbf{Stage 1}\\Tier A (Studio)\\$\text{LR} = 2\mathrm{e}^{-4}$};
\node (s2_std) [below=of s1_std, stage] {\textbf{Stage 2}\\Tier B (Broadcast)\\$\text{LR} = 1\mathrm{e}^{-4}$};
\node (s3_std) [below=of s2_std, stage] {\textbf{Stage 3}\\Tier C+A (Mix)\\$\text{LR} = 1\mathrm{e}^{-5}$};
\draw [arrow] (s1_std) -- (s2_std);
\draw [arrow] (s2_std) -- (s3_std);
\node [above=0.1cm of s1_std, header] {Standard MFT\\(Easy-to-Hard)};
% --- Reverse-MFT (Right Side) ---
\node (s1_rev) [right=0.6cm of s1_std, revstage] {\textbf{Stage 1}\\Tier C (Spontaneous)\\$\text{LR} = 2\mathrm{e}^{-4}$};
\node (s2_rev) [below=of s1_rev, revstage] {\textbf{Stage 2}\\Tier B (Broadcast)\\$\text{LR} = 1\mathrm{e}^{-4}$};
\node (s3_rev) [below=of s2_rev, revstage] {\textbf{Stage 3}\\Tier A+C (Mix)\\$\text{LR} = 1\mathrm{e}^{-5}$};
\draw [arrow] (s1_rev) -- (s2_rev);
\draw [arrow] (s2_rev) -- (s3_rev);
\node [above=0.1cm of s1_rev, header, color=blue!80!black] {R-MFT (Ours)\\(Hard-to-Easy)};
\end{tikzpicture}
\caption{Comparison of training curricula. Both use identical decreasing LR schedules. R-MFT (right) places spontaneous data in the high-LR phase.}
\label{fig:curriculum}
\vspace{-0.6cm}
\end{figure}

% ================================================================
% 2. RELATED WORK
% ================================================================
\section{Related Work}
\label{sec:related}

\textbf{Indic ASR corpora and benchmarks.}
Open-source datasets for Indian languages have expanded rapidly. Kathbath~\cite{kathbath2022} provides large-scale read speech, Shrutilipi~\cite{bhogale2023effectiveness} broadcast news transcriptions, and Indic Voices~\cite{javed2024indicvoices} crowdsourced spontaneous speech. Benchmarks such as Vistaar~\cite{bhogale23_interspeech} evaluate models across multiple domains, analogous to ESB for English~\cite{gandhi2022esb}. Vividh-ASR complements these resources by introducing a complexity-stratified evaluation axis that isolates performance across acoustic difficulty rather than domains.

\textbf{Curriculum learning for ASR.}
Curriculum learning is widely used in ASR training, typically exposing models to progressively noisy data~\cite{tan25b_interspeech,pillai_multistage}. Anti-curriculum and self-paced strategies have been explored in other domains~\cite{jarca2025task}, but their role in adapting large pre-trained speech models remains underexplored. Our work studies how curriculum direction interacts with optimization schedules during fine-tuning.

\textbf{Mechanics of adaptation.}
Adapting multilingual models like Whisper~\cite{radford2023robust} to low-resource languages poses a distinct challenge: learning new phonotactics without degrading the pre-trained acoustic representation. Prior work has analyzed the internal representations of frozen speech models~\cite{pasad2021layerwise,pasad2023comparative}, but the fine-tuning process itself is typically evaluated solely through downstream WER. We bridge this gap by examining how fine-tuning schedules physically reshape internal representations using centered kernel alignment (CKA)~\cite{kornblith2019similarity,merchant2020happens} and singular value decomposition (SVD), linking empirical optimization choices to shifts in model geometry.

% ================================================================
% 3. VIVIDH-ASR BENCHMARK
% ================================================================
\section{The Vividh-ASR Benchmark}
\label{sec:benchmark}

Vividh-ASR\footnote{\url{https://huggingface.co/collections/adalat-ai/vividh-asr}} is a diagnostic benchmark organized by \textit{acoustic and prosodic complexity} rather than by domain. It targets Hindi and Malayalam, representing the Indo-Aryan and Dravidian language families respectively, and aggregates data from Kathbath~\cite{kathbath2022}, Shrutilipi~\cite{bhogale2023effectiveness}, Indic Voices~\cite{javed2024indicvoices}, FLEURS~\cite{fleurs2022arxiv}, and additional publicly available corpora~\cite{baby2016resources}.

\subsection{Tier Definitions}
\label{ssec:tiers}

\begin{itemize}
    \item \textbf{Tier A (Studio):} Scripted, read speech in controlled environments. Clear articulation, standard pronunciation, deliberate pace. Serves as a performance ceiling reference.
    \item \textbf{Tier B (Broadcast):} Read speech from news broadcasts. Clean audio but significantly higher speech rate than Tier~A, testing temporal modeling.
    \item \textbf{Tier C (Spontaneous):} Crowdsourced, unscripted recordings with disfluencies, varying prosody, background noise, and non-professional hardware. The primary bottleneck for real-world Indic ASR.
    \item \textbf{Tier D (Noise):} Tier~A audio augmented with synthetic noise profiles (babble, music, environmental). Held out from training; used exclusively for zero-shot evaluation of acoustic robustness transfer.
\end{itemize}

\subsection{Data Statistics}
\label{ssec:data_stats}

Table~\ref{tab:data_stats} summarizes the corpus. The distribution is intentionally weighted toward Tier~C, reflecting our focus on closing the spontaneous-speech performance gap.

\begin{table}[th]
\caption{Data distribution in hours. The corpus is intentionally weighted toward spontaneous speech (Tier~C). Tier~D is evaluation-only.}
\label{tab:data_stats}
\centering
\scriptsize
\setlength{\tabcolsep}{2pt}
\begin{tabular*}{\columnwidth}{@{\extracolsep{\fill}}l | cc | cc | cc@{}}
\toprule
\textbf{Tier} & \multicolumn{2}{c|}{\textbf{Train (h)}} & \multicolumn{2}{c|}{\textbf{Val (h)}} & \multicolumn{2}{c}{\textbf{Eval (h)}} \\
 & \textbf{Mal} & \textbf{Hi} & \textbf{Mal} & \textbf{Hi} & \textbf{Mal} & \textbf{Hi} \\
\midrule
A (Studio) & 182.2 & 272.1 & 10.2 & 12.0 & 4.28 & 12.50 \\
B (Broadcast) & 200.0 & 1359.91 & 12.6 & 13.2 & 6.33 & 6.82 \\
C (Spontaneous) & 512.5 & 558.65 & 20.1 & 22.1 & 9.62 & 17.43 \\
D (Noise) & -- & -- & -- & -- & 3.08 & 3.00 \\
\midrule
\textbf{Total} & \textbf{894.7} & \textbf{2190.66} & \textbf{42.9} & \textbf{47.3} & \textbf{23.31} & \textbf{39.75} \\
\bottomrule
\end{tabular*}
\end{table}

% ================================================================
% 4. METHODOLOGY
% ================================================================
\section{Methodology}
\label{sec:method}

Standard Whisper fine-tuning relies on conservative learning rates ($1{e}^{-5}$) under the assumption that large updates will destroy the pre-trained priors~\cite{tripathi2025enhancing}. However, when adapting to low-resource languages with complex phonotactics, the model must escape a loss basin shaped by the pretraining distribution. We hypothesize that data scaling alone is insufficient; optimization plasticity (learning rate) and data ordering (curriculum) must be manipulated systematically.

\subsection{Controlled Factorial Design}
\label{ssec:ablations}

To disentangle the effects of optimization plasticity and data ordering, we designed a controlled $2 \times 2$ factorial ablation over two axes:
\begin{enumerate}
\item \textbf{Learning Rate (LR) Timing:} We compare a decreasing schedule ($2\mathrm{e}^{-4} \to 1\mathrm{e}^{-4} \to 1\mathrm{e}^{-5}$) against an increasing schedule ($1\mathrm{e}^{-5} \to 1\mathrm{e}^{-4} \to 2\mathrm{e}^{-4}$). This isolates \textit{when} the model receives large parameter updates.
\item \textbf{Curriculum Direction:} We compare an easy-to-hard ordering (Tier A $\to$ Tier B $\to$ Tier C) against a hard-to-easy ordering (Tier C $\to$ Tier B $\to$ Tier A).
\end{enumerate}
This experimental design allows us to isolate whether acoustic robustness is primarily a function of the data seen, or the plasticity of the model when it sees that data.

\subsection{Reverse Multi-Stage Fine-Tuning (R-MFT)}
\label{ssec:rmft}

Based on the empirical findings of this factorial study (detailed in Section~\ref{sec:results}), we propose Reverse Multi-Stage Fine-Tuning (R-MFT). R-MFT pairs the optimal conditions from our ablation: a high initial learning rate to break out of the pre-trained basin, coupled with a spontaneous-first (hard-to-easy) curriculum. The recipe consists of three stages:

\begin{enumerate}
\item \textbf{Stage 1} (Spontaneous, LR $= 2\mathrm{e}^{-4}$): Tier C data. The highest-plasticity phase encounters the highest-complexity data, explicitly building robustness to disfluencies, noise, and varying prosody.
\item \textbf{Stage 2} (Broadcast, LR $= 1\mathrm{e}^{-4}$): Tier B data. Refines temporal modeling for rapid speech.
\item \textbf{Stage 3} (Consolidation, LR $= 1\mathrm{e}^{-5}$): A 1:1 mixture by duration of Tier A and Tier C. This multi-objective stage acts as a regularizer, recovering any spontaneous performance degraded during Stage 2 while optimizing studio precision.
\end{enumerate}
\begin{table}[th]
\caption{Ablation matrix. Conditions 3--6 form a $2 \times 2$ factorial over LR direction and curriculum direction.}
\label{tab:ablation_design}
\centering
\scriptsize
\setlength{\tabcolsep}{2pt}
\begin{tabular*}{\columnwidth}{@{\extracolsep{\fill}}cl l l@{}}
\toprule
\textbf{\#} & \textbf{Condition} & \textbf{LR Schedule} & \textbf{Curriculum} \\
\midrule
1 & Single-stage, low LR & $1\mathrm{e}^{-5}$ cosine & All tiers mixed \\
2 & Single-stage, high LR & $2\mathrm{e}^{-4}$ cosine & All tiers mixed \\
\midrule
3 & Standard MFT & $2\mathrm{e}^{-4} \to 1\mathrm{e}^{-4} \to 1\mathrm{e}^{-5}$ & A $\to$ B $\to$ C+A \\
4 & \textbf{R-MFT (ours)} & $2\mathrm{e}^{-4} \to 1\mathrm{e}^{-4} \to 1\mathrm{e}^{-5}$ & C $\to$ B $\to$ A+C \\
\midrule
5 & Increasing LR, E$\to$H & $1\mathrm{e}^{-5} \to 1\mathrm{e}^{-4} \to 2\mathrm{e}^{-4}$ & A $\to$ B $\to$ C+A \\
6 & Increasing LR, H$\to$E & $1\mathrm{e}^{-5} \to 1\mathrm{e}^{-4} \to 2\mathrm{e}^{-4}$ & C $\to$ B $\to$ A+C \\
\bottomrule
\end{tabular*}
\end{table}

\subsection{Implementation Details}
\label{ssec:implementation}

We evaluate on Whisper-small and Whisper-medium. All training stages utilize AdamW with weight decay $0.1$. Each stage uses linear warmup for the first 10\% of steps followed by cosine annealing. We train with a batch size of 128 and gradient checkpointing to reduce memory used while training. Each stage trains for a few epochs. Tier D data is strictly held out from all training and validation splits. Models are trained using HuggingFace Transformers on NVIDIA H100 GPUs.

\vspace{-0.2cm}
\section{Results}
\label{sec:results}

\subsection{Learning Rate Effect}
\label{ssec:loss_analysis}

\begin{figure}[h]
\centering
\includegraphics[width=0.8\columnwidth]{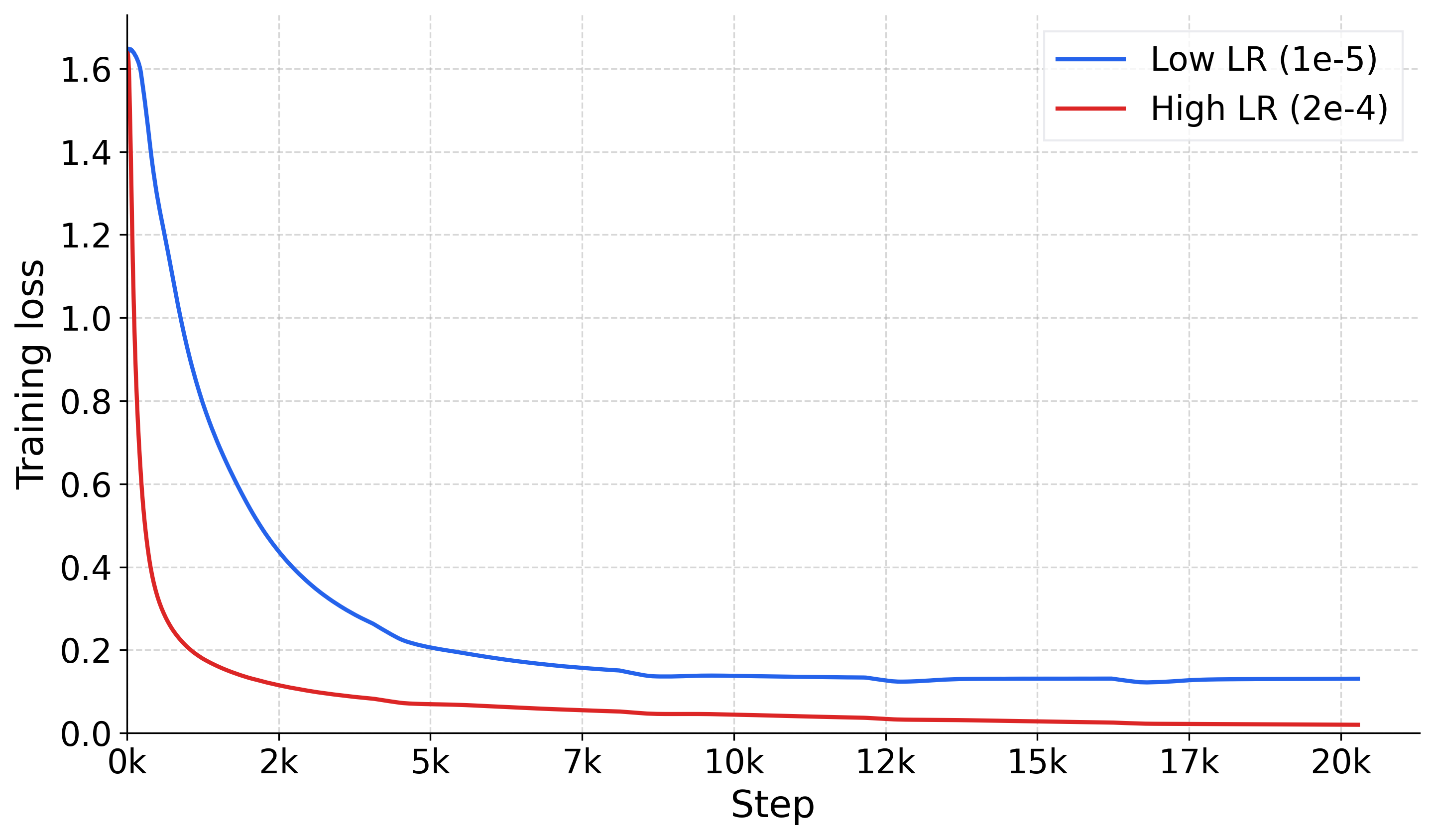}
\caption{Training loss: high LR ($2\mathrm{e}^{-4}$) vs.\ low LR ($1\mathrm{e}^{-5}$) for Malayalam Whisper-medium. The low-LR schedule plateaus prematurely. Identical trends hold for Hindi and Whisper-small.}
\label{fig:loss_curves}
\vspace{-0.4cm}
\end{figure}

Figure~\ref{fig:loss_curves} shows training loss for the Malayalam Whisper-medium model (representative; Hindi and Whisper-small exhibit identical trends). The conservative LR ($1\mathrm{e}{-5}$) plateaus within the first 7K steps at a loss an order of magnitude higher than the $2\mathrm{e}{-4}$ schedule, consistent with the hypothesis that the pre-trained prior creates a deep, narrow basin from which small gradients cannot escape.

\subsection{Overview}
\label{ssec:overview}

Table \ref{tab:main_results} shows both high-LR strategies (R-MFT and Single-stage high LR) vastly outperform the conservative single-stage low-LR baseline (77.79\%/25.25\% global WER). While R-MFT yields the best results for Malayalam (39.36\%), we find that Single-stage high LR is slightly superior for Hindi (16.67\% vs 18.82\% for R-MFT). Both methods significantly exceed the performance of IndicWhisper. This suggests that while the hard-to-easy curriculum is optimal for the more complex phonotactics of Malayalam, simply escaping the low-LR local minima is the primary requirement for Hindi.

% We note that while our training corpus includes Tier C data that the IndicWhisper baseline did not use, the controlled within-method ablations against models trained on identical data distributions constitute our primary evidence for the efficacy of the R-MFT recipe.

\subsection{Effect of Learning Rate Timing}
\label{ssec:lr_timing}

Comparing schedule directions in Table \ref{tab:lr_timing_effect} reveals that LR timing is the primary determinant of performance. In Malayalam, starting with a low LR locks the model into a representational trajectory that later high-LR stages cannot reverse, resulting in a consistent $\sim$13-point penalty across curricula. This locking effect is most severe under R-MFT, where a low-LR start leads to initial catastrophic failure from which the model never fully recovers.

\begin{table}[h]
\centering
\small 
\setlength{\tabcolsep}{3pt} 
\caption{Impact of LR Timing vs. Curriculum Direction on Final Global WER (Malayalam, 769M). Decreasing LR (High $\rightarrow$ Low) consistently avoids the sub-optimal basins that trap increasing-LR schedules.}
\label{tab:lr_timing_effect}
\begin{tabularx}{\columnwidth}{@{}X l l c@{}} 
\toprule
\textbf{Curriculum} & \textbf{Ordering} & \textbf{LR Schedule} & \textbf{WER (\%)} \\ \midrule
Standard MFT & Easy $\rightarrow$ Hard & Dec. (H$\rightarrow$L) & 42.25 \\
Standard MFT & Easy $\rightarrow$ Hard & Inc. (L$\rightarrow$H) & 55.21 \\ \cmidrule(l){2-4} 
\textit{Gap} & & & \textit{+12.96} \\ \midrule
\textbf{R-MFT (Ours)} & Hard $\rightarrow$ Easy & Dec. (H$\rightarrow$L) & \textbf{39.35} \\
R-MFT (Ours) & Hard $\rightarrow$ Easy & Inc. (L$\rightarrow$H) & 51.86 \\ \cmidrule(l){2-4} 
\textit{Gap} & & & \textit{+12.51} \\ \bottomrule
\end{tabularx}
\end{table}

Hindi mirrors this requirement for early plasticity. Regardless of curriculum direction, high-to-low schedules converge to $\sim$18.8\% global WER (improving from 28.49\% initially for standard and 22.71\% for reverse). These results confirm that high-magnitude updates must be applied at the outset to escape the pre-trained prior; delayed high-energy updates are insufficient to exit sub-optimal representational basins formed during a conservative initialization.
\vspace{-0.2cm}
\subsection{Effect of Curriculum Ordering}
\label{ssec:curriculum_effect}

Holding the LR schedule constant isolates the impact of data ordering. In Malayalam, hard-to-easy ordering (R-MFT, 39.35\% WER) consistently outperforms the standard easy-to-hard MFT (42.25\% WER). While this $\sim$3-point gain is smaller than the $\sim$13-point timing effect, it is critical for robustness: exposing the model to the highest-complexity spontaneous data (Tier C) during its initial high-plasticity phase allows the decoder to adapt to disfluencies and non-professional acoustics before refining on cleaner tiers. 

In contrast, Hindi results show convergence to $\sim$18.8\% global WER for both curriculum directions. This suggests that while high-energy initialization is a mandatory requirement for both languages to escape the pre-trained prior, the specific ordering of acoustic complexity provides a specialized advantage for the more challenging phonotactic and prosodic landscape of Malayalam.
\vspace{-0.2cm}
\subsection{Parameter Scale}
\label{ssec:efficiency}

The R-MFT recipe enables remarkable parameter efficiency. Our 244M R-MFT (Small) model achieves 44.41\% (Mal) and 21.41\% (Hi) global WER. This outperforms the much larger 769M Single-stage low-LR baseline by 33.38 and 3.84 absolute points respectively, despite having $1/3$ the parameter capacity. Furthermore, the 244M model exceeds the 769M IndicWhisper baseline in both languages, demonstrating that a "hard-data first" optimization trajectory is more effective than raw model scaling for robust Indic ASR.

\begin{table*}[t]
\caption{WER (\%) across Vividh-ASR tiers for Malayalam (Mal) and Hindi (Hi). Best per-column in \textbf{bold}. $\dagger$: uses less Tier~C training data than our models (see text).}
\label{tab:main_results}
\centering
\small
\begin{tabularx}{\textwidth}{l | c | cc | cc | cc | cc | cc}
\toprule
\textbf{Model} & \textbf{Params} & \multicolumn{2}{c|}{\textbf{Tier A (Studio)}} & \multicolumn{2}{c|}{\textbf{Tier B (Broadcast)}} & \multicolumn{2}{c|}{\textbf{Tier C (Spont.)}} & \multicolumn{2}{c|}{\textbf{Tier D (Noise)}} & \multicolumn{2}{c}{\textbf{Global}} \\
 & & Mal & Hi & Mal & Hi & Mal & Hi & Mal & Hi & Mal & Hi \\
\midrule
IndicWhisper$^\dagger$ & 769M & 33.01 & 16.20 & 33.11 & 11.62 & 66.09 & 39.87 & 48.07 & 14.99 & 48.64 & 25.01 \\
Single-stage, low LR & 769M & 55.68 & 24.01 & 78.64 & 16.22 & 82.37 & 30.62 & 82.17 & 21.60 & 77.79 & 25.25 \\
\midrule

Single-stage, high LR & 769M & \textbf{27.31} & \textbf{12.38} & 30.48 & 11.33 & 50.30 & \textbf{22.99} & 50.78 & \textbf{14.05} & 40.39 & \textbf{16.67} \\
Standard MFT & 769M & 33.56 & 16.41 & 32.75 & 10.91 & 51.03 & 24.91 & 51.51 & 16.25 & 42.25 & 18.81 \\
\textbf{R-MFT (Medium)} & 769M & 31.66 & 16.09 & \textbf{31.66} & \textbf{10.11} & \textbf{46.18} & 24.91 & \textbf{45.73} & 17.27 & \textbf{39.36} & 18.82 \\
% \midrule
% Inc.\ LR, E$\to$H & 769M & XX.X & XX.X & XX.X & XX.X & XX.X & XX.X & XX.X & XX.X & 55.0 & XX.X \\
% Inc.\ LR, H$\to$E & 769M & XX.X & XX.X & XX.X & XX.X & XX.X & XX.X & XX.X & XX.X & 51.0 & XX.X \\
% \midrule
\textbf{R-MFT (Small)} & 244M & 36.49 & 19.16 & 35.05 & 11.49 & 53.74 & 27.34 & 48.04 & 20.97 & 44.41 & 21.41 \\
\bottomrule
\end{tabularx}
\end{table*}

As shown, the R-MFT recipe utilizes its parameter capacity highly efficiently. However, macroscopic WER improvements alone do not explain \textit{why} conservative fine-tuning fails to generalize to complex acoustics, nor how high-LR schedules safely modify the network. To understand the underlying mechanics of this adaptation, we must examine the internal geometry of the models.

% ================================================================
% 6. ANALYSIS
% ================================================================
\vspace{-0.2cm}
\section{Analysis}
\label{sec:analysis}

We hypothesize that successful adaptation to low-resource Indic phonotactics requires a structural asymmetry: learning new linguistic priors in the decoder while preserving the pre-trained encoder's acoustic invariance.

To test this, we trace representational shifts in Whisper-medium on Malayalam---which exhibited the largest performance gap between schedules and thus the clearest signal for representational diagnosis---relative to the base model using four complementary metrics: (i) relative $L_2$ weight displacement ($\Delta\theta$) to measure optimization plasticity, (ii) Centered Kernel Alignment (CKA) to verify preservation of activation geometry, {(iii) exact Optimal Transport (Wasserstein-1 EMD)~\cite{EMD_1998} to capture distribution drift, and (iv) Singular Value Decomposition (SVD) to detect structural fragmentation.

\subsection{Encoder vs. Decoder Adaptation}

Table~\ref{tab:activation_analysis} reveals a critical structural asymmetry that distinguishes our high-LR methods. Successful adaptation is characterized by substantial parameter displacement concentrated almost exclusively in the decoder (mean $\Delta\theta = 0.122$ for R-MFT). This is mirrored by a significant shift in decoder activation distributions (EMD $= 0.069$), indicating that the model is actively re-mapping the linguistic prior.

Crucially, the encoder’s pre-trained geometry remains invariant. For both Baseline-High and R-MFT, Encoder CKA remains perfect ($1.000$) with near-zero EMD. This demonstrates that an aggressive initial step size ($2\mathrm{e}^{-4}$) does not destroy the encoder; rather, it provides the necessary gradient energy for the decoder to escape the pre-trained basin while the encoder remains anchored to its robust acoustic features. Conversely, the conservative low-LR baseline fails to generate sufficient displacement ($\Delta\theta \approx 0.01$) to achieve any meaningful adaptation.

\begin{table}[th]
\caption{Layer-wise representational shifts relative to the base Whisper-medium model.}
\label{tab:activation_analysis}
\centering
\footnotesize
\setlength{\tabcolsep}{2pt}
\begin{tabular*}{\columnwidth}{@{\extracolsep{\fill}}l | cc | cc | cc@{}}
\toprule
\textbf{Model vs.\ Base} & \multicolumn{2}{c|}{\textbf{CKA}} & \multicolumn{2}{c|}{\textbf{EMD}} & \multicolumn{2}{c}{\textbf{$\Delta\theta$ ($L_2$)}} \\
 & \textbf{Enc} & \textbf{Dec} & \textbf{Enc} & \textbf{Dec} & \textbf{Enc} & \textbf{Dec} \\
\midrule
IndicWhisper & 0.775 & 0.993 & 0.605 & 0.467 & 0.025 & 0.016 \\
Baseline-Low LR & 1.000 & 0.999 & 0.000 & 0.168 & 0.010 & 0.018 \\
Baseline-High LR & 1.000 & 0.999 & 0.000 & 0.100 & 0.080 & 0.123 \\
\midrule
R-MFT Stage 1 & 1.000 & 0.999 & 0.000 & 0.072 & 0.073 & 0.112 \\
R-MFT Final & \textbf{1.000} & \textbf{0.999} & \textbf{0.000} & \textbf{0.069} & \textbf{0.076} & \textbf{0.122} \\
\bottomrule
\end{tabular*}
\end{table}
\vspace{-0.4cm}
\subsection{Spectral Signatures of Studio-Bias}

IndicWhisper, fine-tuned predominantly on read and broadcast speech, exhibits a fundamentally different structural signature. Despite lower overall weight displacement than R-MFT ($\Delta\theta = 0.025$), it severely disrupts the pre-trained encoder's geometry, dropping CKA to $0.775$.

We verify this shift via rank expansion (Table~\ref{tab:svd_analysis}). While the base model and R-MFT maintain a compact Encoder Effective Rank ($\zeta \approx 14$), IndicWhisper expands this to $\zeta=25$. Although differing training distributions preclude a strictly causal comparison, this expansion suggests that clean-acoustic adaptation overwrites the encoder's robust feature space with studio-specific nuances. This ``over-parameterization'' correlates with the steep degradation observed on spontaneous speech (66.09\% WER), as the model loses the generalized acoustic invariance of the pre-trained prior.

In contrast, our R-MFT schedule---driven by a hard-to-easy curriculum and high initial learning rates---achieves superior adaptation entirely through decoder realignment, leaving the foundational acoustic robustness of the encoder intact.

\begin{table}[th]
\caption{SVD analysis of encoder and decoder activations.}
\label{tab:svd_analysis}
\centering
\footnotesize
\setlength{\tabcolsep}{4pt}
\begin{tabular*}{\columnwidth}{@{\extracolsep{\fill}}l | cc | cc@{}}
\toprule
\textbf{Model} & \multicolumn{2}{c|}{\textbf{Effective Rank}} & \multicolumn{2}{c}{\textbf{Tail $S_{-1}$}} \\
 & \textbf{Enc} & \textbf{Dec} & \textbf{Enc} & \textbf{Dec} \\
\midrule
Whisper base & 14 & 9 & .0134 & .0018 \\
IndicWhisper & 25 & 11 & .0443 & .0045 \\
Baseline-High LR & 14 & 9 & .0134 & .0019 \\
R-MFT Final & 14 & 9 & .0134 & .0017 \\
\bottomrule
\end{tabular*}
\end{table}

\vspace{-0.2cm}
\section{Conclusion}
\label{sec:conclusion}

We introduced Vividh-ASR, a complexity-tiered benchmark designed to diagnose \textit{studio-bias} in Indic speech recognition. Through a $2 \times 2$ factorial study on Hindi and Malayalam, we showed that aggressive early learning rates are the dominant factor, yielding $\sim12$ absolute WER points; later high-LR stages cannot recover performance lost to a conservative, low-LR initialization. Curriculum direction contributes a smaller, language-dependent gain: within a multi-stage fine-tuning setting, a reverse hard-to-easy curriculum (R-MFT) outperforms standard MFT, most notably on spontaneous Malayalam speech.

Motivated by these dynamics, we proposed R-MFT, enabling a 244M Whisper model to match or exceed conventionally fine-tuned 769M counterparts. CKA and SVD analyses confirm this strategy adapts the decoder to target-language phonotactics while preserving the encoder's robust acoustic geometry. Future work will extend Vividh-ASR to additional languages and investigate if these optimization dynamics generalize beyond Whisper to self-supervised and Conformer-based models \cite{gulati2020conformer}. Finally, given our mechanistic findings regarding encoder invariance, we specifically aim to explore the efficacy of selective encoder freezing as a regularization strategy to further mitigate studio-bias.

\section{Generative AI Use Disclosure}

The authors utilized large language model (LLM) tools, specifically Gemini 2.5 Pro, to assist in the linguistic refinement and technical polishing of the manuscript. All final content was reviewed, verified, and approved by the authors, who take full responsibility for the integrity of the research and its presentation.

\bibliographystyle{IEEEtran}
\bibliography{mybib}

@inproceedings{kirkpatrick2017ewc,
  title={Overcoming catastrophic forgetting in neural networks},
  author={Kirkpatrick, James et al.},
  booktitle={PNAS},
  year={2017}
}

@inproceedings{conneau2021xlsr,
  title={Unsupervised Cross-Lingual Representation Learning for Speech Recognition},
  author={Conneau, Alexis and Baevski, Alexei and Collobert, Ronan and Mohamed, Abdelrahman and Auli, Michael},
  booktitle={Interspeech},
  year={2021}
}

@inproceedings{bengio2009curriculum,
  title={Curriculum Learning},
  author={Bengio, Yoshua and Louradour, J{\'e}r{\^o}me and Collobert, Ronan and Weston, Jason},
  booktitle={ICML},
  year={2009}
}

@inproceedings{gulati2020conformer,
  title={Conformer: Convolution-augmented Transformer for Speech Recognition},
  author={Gulati, Anmol and Qin, James and Chiu, Chung-Cheng and Parmar, Niki and Zhang, Yu and Yu, Jiahui and Han, Wei and Wang, Shibo and Zhang, Zhengdong and Wu, Yonghui and Pang, Ruoming},
  booktitle={Interspeech},
  pages={5036--5040},
  year={2020}
}

@inproceedings{pasad2021layerwise,
  title={Layer-wise analysis of a self-supervised speech representation model},
  author={Pasad, Ankita and Chou, Ju-Chieh and Livescu, Karen},
  booktitle={IEEE Automatic Speech Recognition and Understanding Workshop (ASRU)},
  pages={914--921},
  year={2021},
  organization={IEEE}
}

@inproceedings{pasad2023comparative,
  title={Comparative layer-wise analysis of self-supervised speech models},
  author={Pasad, Ankita and Shi, Bowen and Livescu, Karen},
  booktitle={IEEE International Conference on Acoustics, Speech and Signal Processing (ICASSP)},
  pages={1--5},
  year={2023},
  organization={IEEE}
}

@inproceedings{kornblith2019similarity,
  title={Similarity of neural network representations revisited},
  author={Kornblith, Simon and Norouzi, Mohammad and Lee, Honglak and Hinton, Geoffrey},
  booktitle={International Conference on Machine Learning (ICML)},
  pages={3519--3529},
  year={2019},
  organization={PMLR}
}

@inproceedings{merchant2020happens,
  title={What Happens To {BERT} Embeddings During Fine-tuning?},
  author={Merchant, Amil and Rahimtoroghi, Elahe and Pavlick, Ellie and Tenney, Ian},
  booktitle={Proceedings of the Third BlackboxNLP Workshop on Analyzing and Interpreting Neural Networks for NLP},
  pages={33--44},
  year={2020}
}

@article{jarca2025task,
  title={Task-Informed Anti-Curriculum by Masking Improves Downstream Performance on Text},
  author={Jarca, Andrei and Croitoru, Florinel Alin and Ionescu, Radu Tudor},
  journal={arXiv preprint arXiv:2502.12953},
  year={2025},
  note={Accepted at ACL 2025}
}

@article{gandhi2022esb,
  title={Esb: A benchmark for multi-domain end-to-end speech recognition},
  author={Gandhi, Sanchit and Von Platen, Patrick and Rush, Alexander M},
  journal={arXiv preprint arXiv:2210.13352},
  year={2022}
}

@article{fleurs2022arxiv,
  title = {FLEURS: Few-shot Learning Evaluation of Universal Representations of Speech},
  author = {Conneau, Alexis and Ma, Min and Khanuja, Simran and Zhang, Yu and Axelrod, Vera and Dalmia, Siddharth and Riesa, Jason and Rivera, Clara and Bapna, Ankur},
  journal={arXiv preprint arXiv:2205.12446},
  url = {https://arxiv.org/abs/2205.12446},
  year = {2022},
}

@inproceedings{tan25b_interspeech,
  title     = {CBA-Whisper: Curriculum Learning-Based AdaLoRA Fine-Tuning on Whisper for Low-Resource Dysarthric Speech Recognition},
  author    = {Tianyi Tan and Xinan Chen and Xiaohuai Le and Wenzhi Fan and Xianjun Xia and Chuanzeng Huang and Jing Lu},
  year      = {2025},
  booktitle = {Interspeech 2025},
  pages     = {3309--3313},
  doi       = {10.21437/Interspeech.2025-1705},
  issn      = {2958-1796},
}

@misc{lodagala_whisper_finetune_hyperparams,
  author       = {Lodagala, Vasista},
  title        = {whisper-finetune: Hyperparameter Tuning},
  year         = {2024},
  howpublished = {\url{https://github.com/vasistalodagala/whisper-finetune#hyperparameter-tuning}},
  note         = {GitHub repository, accessed 2026-03-05}
}

@inproceedings{tripathi2025enhancing,
  title={Enhancing whisper’s accuracy and speed for indian languages through prompt-tuning and tokenization},
  author={Tripathi, Kumud and Gothi, Raj and Wasnik, Pankaj},
  booktitle={ICASSP 2025-2025 IEEE International Conference on Acoustics, Speech and Signal Processing (ICASSP)},
  pages={1--5},
  year={2025},
  organization={IEEE}
}

@inproceedings{radford2023robust,
  title={Robust speech recognition via large-scale weak supervision},
  author={Radford, Alec and Kim, Jong Wook and Xu, Tao and Brockman, Greg and McLeavey, Christine and Sutskever, Ilya},
  booktitle={International conference on machine learning},
  pages={28492--28518},
  year={2023},
  organization={PMLR}
}

@inproceedings{bhogale23_interspeech,
  title     = {{Vistaar: Diverse Benchmarks and Training Sets for Indian Language ASR}},
  author    = {Kaushal Bhogale and Sai Sundaresan and Abhigyan Raman and Tahir Javed and Mitesh M. Khapra and Pratyush Kumar},
  year      = {2023},
  booktitle = {{Interspeech 2023}},
  pages     = {4384--4388},
  doi       = {10.21437/Interspeech.2023-2589},
  issn      = {2958-1796},
}

@inproceedings{javed2024indicvoices,
  title={Indicvoices: Towards building an inclusive multilingual speech dataset for indian languages},
  author={Javed, Tahir and Nawale, Janki and George, Eldho and Joshi, Sakshi and Bhogale, Kaushal and Mehendale, Deovrat and Sethi, Ishvinder and Ananthanarayanan, Aparna and Faquih, Hafsah and Palit, Pratiti and others},
  booktitle={Findings of the Association for Computational Linguistics: ACL 2024},
  pages={10740--10782},
  year={2024}
}

@inproceedings{bhogale2023effectiveness,
  title={Effectiveness of mining audio and text pairs from public data for improving ASR systems for low-resource languages},
  author={Bhogale, Kaushal and Raman, Abhigyan and Javed, Tahir and Doddapaneni, Sumanth and Kunchukuttan, Anoop and Kumar, Pratyush and Khapra, Mitesh M},
  booktitle={Icassp 2023-2023 ieee international conference on acoustics, speech and signal processing (icassp)},
  pages={1--5},
  year={2023},
  organization={IEEE}
}

@inproceedings{baby2016resources,
  title        = {Resources for {I}ndian languages},
  author       = {Baby, Arun and Thomas, Anju Leela and Nishanthi, NL and TTS Consortium and others},
  booktitle    = {Proceedings of Text, Speech and Dialogue},
  year         = {2016},
  organization = {CBBLR Workshop}
}

@misc{kathbath2022,
  doi = {10.48550/ARXIV.2208.11761},
  url = {https://arxiv.org/abs/2208.11761},
  author = {Javed, Tahir and Bhogale, Kaushal Santosh and Raman, Abhigyan and Kunchukuttan, Anoop and Kumar, Pratyush and Khapra, Mitesh M.},
  title = {IndicSUPERB: A Speech Processing Universal Performance Benchmark for Indian languages},
  publisher = {arXiv},
  year = {2022},
  copyright = {arXiv.org perpetual, non-exclusive license}
}

@INPROCEEDINGS{EMD_1998,
  author={Rubner, Y. and Tomasi, C. and Guibas, L.J.},
  booktitle={Sixth International Conference on Computer Vision (IEEE Cat. No.98CH36271)}, 
  title={A metric for distributions with applications to image databases}, 
  year={1998},
  volume={},
  number={},
  pages={59-66},
  doi={10.1109/ICCV.1998.710701}}

@article{pillai_multistage,
author = {Pillai, Leena G and Manohar, Kavya and Raju, Basil and Sherly, Elizabeth},
title = {Multistage Fine-tuning Strategies for Automatic Speech Recognition in Low-resource Languages},
year = {2026},
publisher = {Association for Computing Machinery},
address = {New York, NY, USA},
issn = {2375-4699},
url = {https://doi.org/10.1145/3813800},
doi = {10.1145/3813800},
journal = {ACM Trans. Asian Low-Resour. Lang. Inf. Process.},
month = may
}

@inproceedings{nethil25_interspeech,
  title     = {{Scalable Offline ASR for Command-Style Dictation in Courtrooms}},
  author    = {Kumarmanas Nethil and Vaibhav Mishra and Kriti Anandan and Kavya Manohar},
  year      = {2025},
  booktitle = {{Interspeech 2025}},
  pages     = {308--309},
  issn      = {2958-1796},
}

\end{document}